\documentclass[conference]{IEEEtran}
\IEEEoverridecommandlockouts
\UseRawInputEncoding
\usepackage{cite}
\usepackage{amsmath,amssymb,amsfonts}
\usepackage{algorithmic}
\usepackage{graphicx}
\usepackage{textcomp}
\usepackage{xcolor}
\usepackage{longtable}
\usepackage{tabularx}
\usepackage{fancyhdr}
\usepackage[utf8]{inputenc} % allow utf-8 input
\usepackage{gensymb}
\usepackage{enumitem}
\usepackage{lineno}
\usepackage{float} 
\usepackage{multirow}
\usepackage{longtable}
\usepackage{booktabs}
\usepackage{array}
\usepackage{amsmath,array,graphicx}
\usepackage{caption,subcaption}
\usepackage{nameref}
\usepackage{amsmath,amssymb}
\usepackage{gensymb}
\usepackage{tabularray}
\usepackage{adjustbox}
\usepackage{multirow}
\usepackage[pagebackref=true,breaklinks=true,letterpaper=false,colorlinks,bookmarks=false]{hyperref}
\usepackage{orcidlink}

\def\BibTeX{{\rm B\kern-.05em{\sc i\kern-.025em b}\kern-.08em
    T\kern-.1667em\lower.7ex\hbox{E}\kern-.125emX}}

\begin{document}
\newcommand{\BB}[1]{{\color{black}{#1}}}
\newcommand{\RR}[1]{{\color{red}{#1}}}
\newcommand{\GR}[1]{{\color{green}{#1}}}

\title{Accurate and Efficient Urban Street Tree Inventory with Deep Learning on Mobile Phone Imagery}

\author{\IEEEauthorblockN{1\textsuperscript{st} Asim Khan$^*$}
\IEEEauthorblockA{\textit{Department of Mechanical and Nuclear Engineering} \\
\textit{Khalifa University}\\
Abu Dhabi, United Arab Emirates (UAE) \\
asim.khan@ku.ac.ae \orcidlink {0000-0003-0543-3350}}
\and
\IEEEauthorblockN{2\textsuperscript{nd} Umair Nawaz$^*$}
\IEEEauthorblockA{\textit{Department of Electrical Engineering} \\
\textit{Namal University}\\
Mianwali, Pakistan \\
1601005@namal.edu.pk}
\and
\IEEEauthorblockN{3\textsuperscript{rd} Anwaar Ulhaq}
\IEEEauthorblockA{\textit{School of Engineering and Technology, }
\textit{Central Queensland University}\\
Sydney, Australia \\
aulhaq@csu.edu.au}
\and
\IEEEauthorblockN{4\textsuperscript{th} Iqbal Gondal}
\IEEEauthorblockA{\textit{School of Computing Technologies} \\
\textit{Royal Melbourne Institute of Technology (RMIT) University}\\
%\textit{RMIT University}
Melbourne, Australia \\
iqbal.gondal@rmit.edu.au}
\and
\IEEEauthorblockN{5\textsuperscript{th} Sajid Javed}
\IEEEauthorblockA{\textit{Department of Electrical Engineering and Computer Science} \\
\textit{Khalifa University}\\
Abu Dhabi, United Arab Emirates (UAE) \\
sajid.javed@ku.ac.ae}
% \and
}

\maketitle

\begin{abstract}    %The abstract should contain about 100 to 150 words.
\textbf{Deforestation, a major contributor to climate change, poses detrimental consequences such as agricultural sector disruption, global warming, flash floods, and landslides. Conventional approaches to urban street tree inventory suffer from inaccuracies and necessitate specialised equipment. To overcome these challenges, this paper proposes an innovative method that leverages deep learning techniques and mobile phone imaging for urban street tree inventory. Our approach utilises a pair of images captured by smartphone cameras to accurately segment tree trunks and compute the diameter at breast height (DBH). Compared to traditional methods, our approach exhibits several advantages, including superior accuracy, reduced dependency on specialised equipment, and applicability in hard-to-reach areas. We evaluated our method on a comprehensive dataset of 400 trees and achieved a DBH estimation accuracy with an error rate of less than 2.5\%. Our method holds significant potential for substantially improving forest management practices. By enhancing the accuracy and efficiency of tree inventory, our model empowers urban management to mitigate the adverse effects of deforestation and climate change.}
\end{abstract}


\def\thefootnote{*}\footnotetext{These authors contributed equally to this work.}
%------------------------------------------------------

\section{Introduction}
\label{sec:intro}

Standing tall with their verdant foliage, trees are vital pillars in maintaining the delicate balance of our ecosystem.  Additionally, trees provide sheltering shade, purify water sources, and contribute to the overall well-being of communities. The diameter at breast height (DBH) holds immense significance as a pivotal indicator for conducting ecological investigations and studying forest resources.  Consequently, this parameter's accurate and efficient acquisition has long been a focal point for investigators and researchers in the field\cite{mokrovs2018evaluation,rs12142238}. The task of forest inventory management includes the calculation of diameter at breast height (DBH), which is quite a tedious and lengthy procedure, as foresters must do this task manually due to the unavailability of proper equipment, which also introduces human error. Improving the efficiency and accuracy of DBH data gathering is a critical research topic in forest ecology and forest resource digitalization \cite{li2023development, zhou2019estimation,f10090778}. This study focused on estimating the diameter at breast height (DBH) of three tree species commonly found along the streets of Dubai, United Arab Emirates: Phoenix dactylifera (Date species), Vachellia nilotica, and Ziziphus mauritiana.

This paper introduces an efficient and cost-effective approach to segment tree trunks and accurately determine the diameter at breast height (DBH) by leveraging the power of standard Android mobile phones. With a focus on user-friendliness and affordability, our proposed method offers numerous advantages over existing techniques, eliminating the need for specialized equipment and extensive training. By streamlining the process into three essential steps—capturing consecutive images, segmenting the trunk, and calculating the DBH—our method provides a more accessible and cost-effective solution for conducting street tree inventory.

The paper encompasses two distinct yet interrelated objectives. Firstly, we aim to estimate the DBH using a straightforward and economically viable method, enabling efficient and accurate measurements. Secondly, we seek to establish a comprehensive dataset that can serve as a foundation for further research, enabling the development of more advanced and universally applicable models for predicting DBH. By successfully achieving these objectives, our study strives to push the boundaries of DBH estimation methods, elevating the precision and applicability of predictive models in this domain.

Through this paper, we anticipate significant advancements in DBH estimation techniques, leading to improved accuracy and enhanced generalizability of predictive models. By leveraging the convenience and ubiquity of mobile phone imagery, we unlock new possibilities for forest inventory management and ecological studies. Our approach simplifies the process and lays the groundwork for future advancements in this field, fostering innovation and progress in tree measurement techniques.

The paper is organized as follows to comprehensively explore the topic. Section 2 presents an overview of the relevant literature on diameter at breast height (DBH) estimation. Section 3 introduces our proposed approach, which leverages deep learning techniques for accurate DBH calculation. In Section 4, we delve into the specifics of the dataset used and describe the annotation procedure employed. The details of our model training and the application specifics are elucidated in Section 5. Section 6 presents the noteworthy findings and results obtained through our method. Finally, in Section 7, we conclude the paper by summarizing our study's key insights and contributions.

 %%%%%%%%%%%%%%%%%%%%%%%%%%%%%%%%%%%%%%%%%%%%%%%%%%%%%%%%%%%%%%%%
 
\section {Related Work}
The advancement of remote sensing technology and machine vision has led to the emergence of non-contact measurement methods. Artificial neural networks (ANNs) are powerful tools for solving nonlinear issues like forecasting forest development. Many scholars and academics have incorporated ANN models into the field of forestry for the dynamic monitoring of forest resources and the modelling of the forest development process to establish more accurate ways for estimating forest growth \cite{diamantopoulou2005artificial, ozccelik2010estimating, leite2011estimation}.

Hamra et al.,\cite{hamraz2019deep}  used airborne LiDAR data to create 3D point clouds, which they use along with deep learning models for the Coniferous/Deciduous classification of individual trees. They have used discrete representations using leaf-on and leaf-off LiDAR data and an ensemble of convolution neural networks (CNN). Although LiDAR-based systems provide excellent results, they are expensive and require expertise to acquire and process 3D point cloud data. To solve these issues \cite{piermattei2019terrestrial,liang2014use} perform a comparative work between LiDAR and terrestrial photogrammetry, which is based on creating 3D point cloud from camera images along with structure from motion (SfM) and dense matching algorithm, the author reports that the error for terrestrial photogrammetry is higher but given the low cost and requirement of expertise, it is a viable solution.

The application of ultrasound technology in tree measurement involves the calculation of tree distance and the extraction of diameter at breast height (DBH) through the analysis of acoustic wave propagation and time differences. However, implementing ultrasonic methods in field measurements is challenging due to the requirement of long-range measurement systems, which employ high-power transmitters and cumbersome instruments \cite{ahn2008practical,KUMAR2017186}.

Machine vision technology, which leverages image pixel information and camera imaging principles, is a valuable tool for estimating distances and acquiring object sizes. Notably, this approach offers distinct advantages, including the abundance of image information and its cost-effectiveness \cite{sun2017laser, 10.1007/s10586-017-1356-8}. Within the realm of machine vision measurement, both monocular and binocular vision measurements play crucial roles \cite{ROYDEN20167, rs11171990}. Traditionally, early techniques for extracting image information primarily relied on the binocular stereo vision principle and camera motion information, often necessitating the acquisition of multiple images to extract depth information \cite{LI20214325, 9398929, ZHANG2020107260,Yang2017VisionSO}. In contrast, monocular methods offer distinct advantages, as they do not require stringent hardware conditions during image acquisition and facilitate convenient device integration.

In our cost-effective and easy solution, we have opted to use a monocular camera-based approach. We have designed our approach to work with off-the-shelf smartphone cameras since smartphone accessibility and its use have been rising steadily in the country and would not require users to be trained to operate it.

%%%%%%%%%%%%%%%%%%%%%%%%%%%%%%%%%%%%%%%%%%%%%%%%%%%%%%%%%%%%%%%%

\section{The  Proposed Method}
\label{sec:methods}
This paper introduces an innovative method for calculating the diameter at breast height (DBH) by utilizing a deep learning-based algorithm called SegFormer \cite{NEURIPS2021_64f1f27b}.

Our approach capitalizes on the availability of images captured using standard Android mobile phones, enabling the segmentation of tree trunks and accurate DBH calculation. Our proposed method offers numerous advantages by eliminating the reliance on specialized equipment and extensive training, making it a more accessible and cost-effective solution for conducting street tree inventory. The process involves capturing two consecutive images of a tree trunk, segmenting the trunk from these images, and utilizing the segmented images to calculate the DBH.

To estimate the DBH, we utilize the pin camera principle, which relies on projecting a tree trunk onto a mobile camera sensor. Equation (\ref{object_hieght_equation}) illustrates this principle where the height of the tree is calculated:
\begin{equation}
\label{object_hieght_equation}
H = \frac{x \times y}{z}
\end{equation}

In Equation \ref{object_hieght_equation}, H represents the real object height in feet, $x$ denotes the distance to the object, $y$ represents the object height on the camera sensor, and $z$ corresponds to the focal length of the camera.

To utilize this equation, two values must be computed: the distance between the camera and the object (x) and the size of the tree trunk's projection on the camera sensor (y). The distance is determined using a binary image approach that takes advantage of the variance in object size projection onto a camera lens based on the object's distance from the camera. This approach involves capturing two consecutive images: a far image and a close image. The close image is taken after moving the camera a certain distance closer to the tree from the far image's location. The methodology behind this approach is inspired by \cite{equation_distance}. 

\begin{figure}
\centering
\includegraphics[scale=0.35]{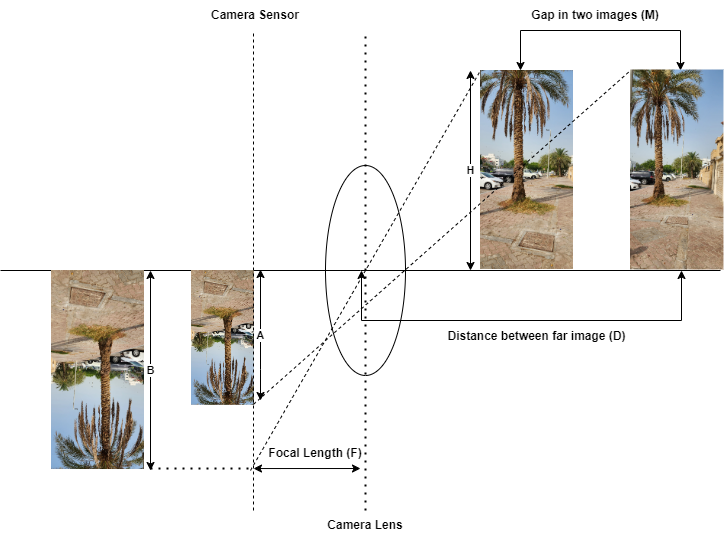}
\caption{Distance Calculation through Multi-Image Acquisition Approach and Visual Tree Projection Observation on Camera Sensor} \label{distance_to_tree_fig}
\end{figure}

Figure~\ref{distance_to_tree_fig} shows this approach visually, where the projection of the tree is reflected on the camera sensor.

To ensure precise calculations, it is imperative to segment equal portions of the tree trunks from the far and close images. To accomplish this task, we employ the SegFormer \cite{NEURIPS2021_64f1f27b} model, which is explicitly fine-tuned for our dataset, enabling accurate tree trunk segmentation. Additional details regarding the dataset collected can be found in the upcoming section of the paper. SegFormer is a state-of-the-art semantic segmentation model that has gained significant attention in the computer vision domain. It is a transformer-based architecture that combines the power of transformers with the effectiveness of convolutional neural networks (CNNs) for pixel-level segmentation tasks. SegFormer achieves impressive results by leveraging the self-attention mechanism of transformers to capture long-range dependencies in images, enabling it to understand the context and relationships between pixels effectively. By incorporating CNNs into the architecture, SegFormer captures local spatial information and thus further enhances its segmentation performance. With its ability to handle complex scenes and produce accurate segmentation masks, SegFormer has emerged as a promising model for various applications, including object detection, image segmentation, and scene understanding \cite{NEURIPS2021_64f1f27b}.

Once the segmentation masks have been obtained, it is essential to isolate equivalent portions of the masks. To achieve this, we utilize a warping approach that aligns the far image with the close image. This warping process incorporates an estimated transformation to ensure optimal alignment, particularly in terms of the shared region of the tree. Subsequently, we extract the region of interest from the far image, acquiring the common area present in both images.

To ascertain the object's height on the sensor, an Android application is utilized to extract essential parameters such as the width, height, and focal length of the sensor, as well as the pixel dimensions of the resulting image. Subsequently, a scaling factor is computed by comparing the pixel height of the segmented trunk in the image with the original image height of the tree. The precise height of the object on the sensor is calculated by multiplying this scaling factor by the sensor height. The real-world height of the tree trunk is calculated by substituting this value into Equation (\ref{object_hieght_equation}). This height measurement is a basis for calculating the measurements in physical units such as feet. This value is obtained by dividing the actual object height by the number of pixels present in the original mask of the tree trunk in the far image. This value represents the conversion factor from pixel units to real-world distance, i.e. feet, and is referred to as the DF in Equation~\ref{DBH-equation}. 

To estimate the diameter at breast height (DBH), we convert the breast height value (i.e. 4.5 feet) to its corresponding pixel height using the previous scaling factor. This pixel height serves as a reference for extracting the diameter measurement from the original segmentation mask of the tree trunk, specifically from the far image.

Finally, to obtain the DBH in real-world distance units, we multiply the extracted diameter in pixels by the distance factor as shown in Equation~\ref{DBH-equation}, which can also be termed a PDF formula. This conversion process allows us to translate the pixel-based diameter measurement to its equivalent physical measurement, i.e. cm.

\begin{equation}
\label{DBH-equation}
\centering
DBH = P \times DF
\end{equation}

Here in Equation~\ref{DBH-equation}, P is the number of pixels at breast height, and DF is the distance factor used to convert the pixels into centimeters.

\begin{figure*}[t]
\centering
\includegraphics[scale=0.35]{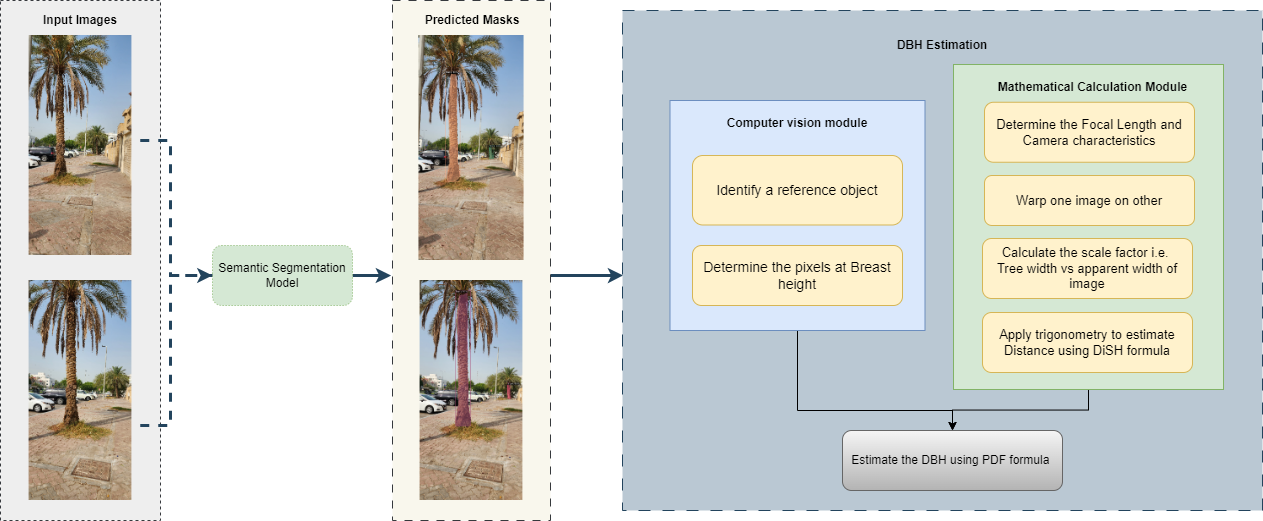}
\caption{A Workflow Diagram for Tree Trunk Diameter Estimation}
\label{workflow_figure1}
\end{figure*}

\begin{figure*}[t]
\centering
\includegraphics[scale=0.35]{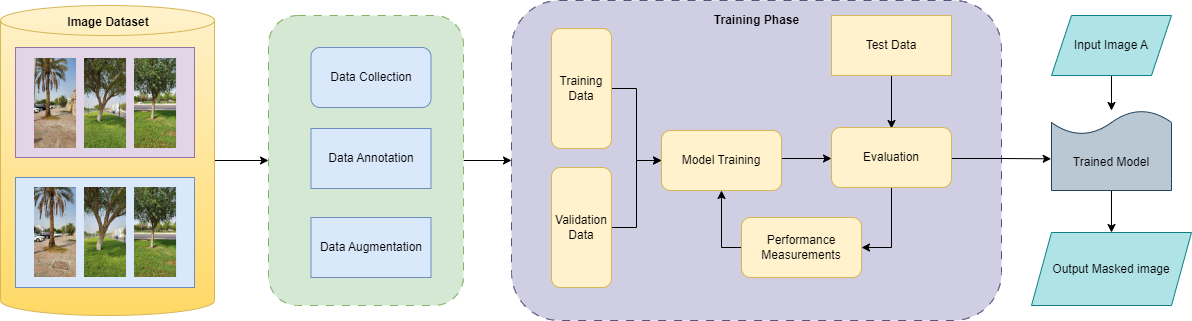}
\caption{A Block Diagram for Tree Trunk Segmentation and Diameter Estimation}
\label{workflow_figure2}
\end{figure*}

The entire workflow of the application is illustrated in Figures \ref{workflow_figure1} and \ref{workflow_figure2}. Here, a pair of input images is provided by the Android application, which is processed through the SegFormer model and thus, the segmented masks are extracted. Furthermore, these masks are passed to the DBH estimation algorithm, where the DBH is estimated based on mathematical and photogrammetric principles.

%_____________________________________________________________

\begin{figure}[t]
\begin{center}
\includegraphics[scale=0.30]{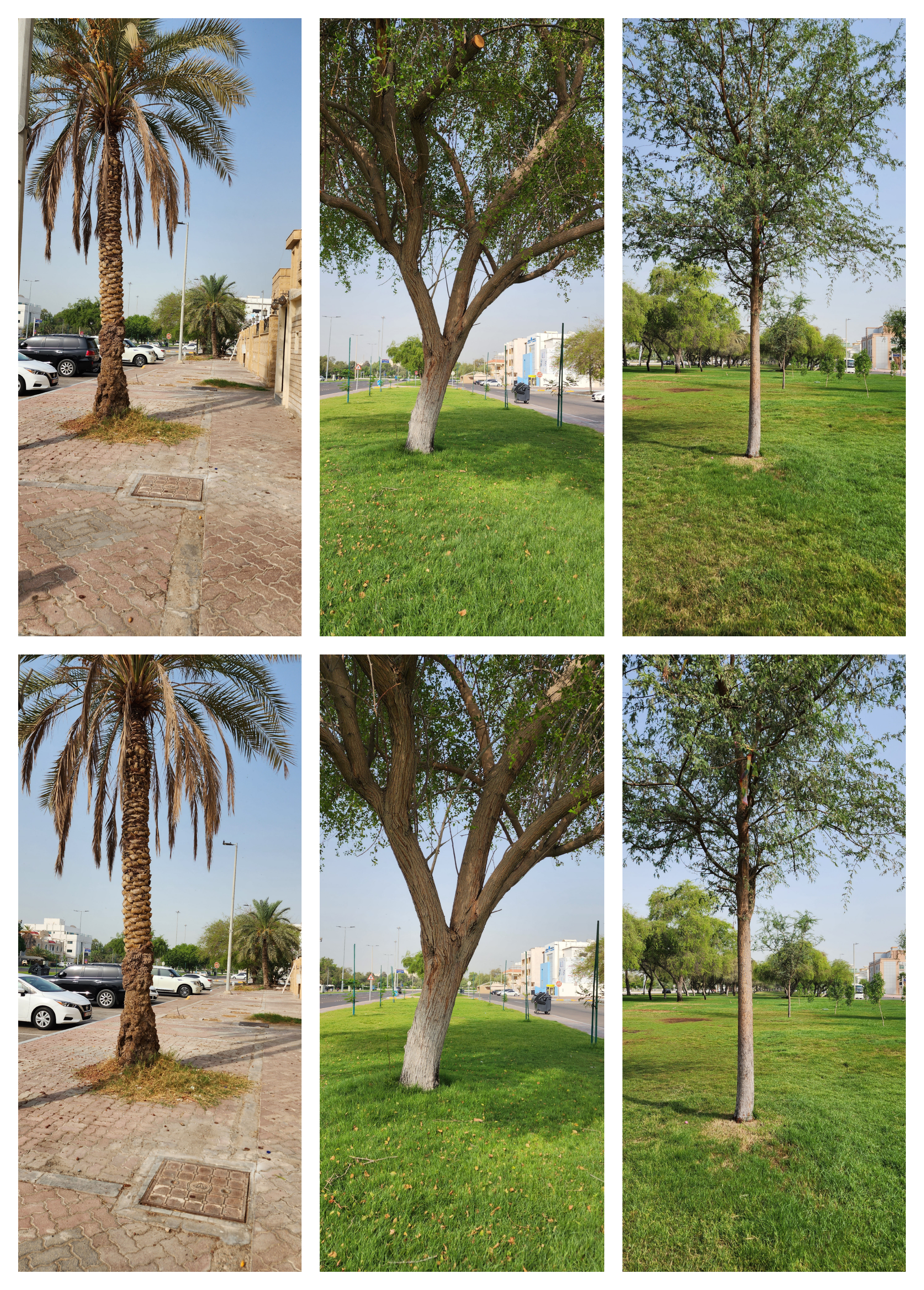}
\end{center}
\caption{The dataset includes sample images captured at different distances. The images in the first row, namely Phoenix dactylifera, Ziziphus mauritiana, and Vachellia nilotica, were taken at a distance of 20 feet. On the other hand, the images in the second row, also consisting of Phoenix dactylifera, Ziziphus mauritiana, and Vachellia nilotica, were captured at a distance of 15 feet.}
\label{sample_dataset_images}
\end{figure}

%%%%%%%%%%%%%%%%%%%%%%%%%%%%%%%%%%%%%%%%%%%%%%%%%%%%%

\section{Experimental Setup}
\label{sec:experiment}
%\vspace{-2mm}
\subsection{Dataset preparation}
The data collection for our study was conducted in the urban streets of Dubai, UAE. Most of the trees in this area belong to the Date species (Phoenix dactylifera), along with other species such as Vachellia nilotica and Ziziphus mauritiana.

As part of our academic paper, we collected images explicitly focusing on these tree species. This data collection aimed to obtain a representative sample of the trees found in urban street environments in the UAE. By capturing images of these specific species, we aimed to analyze and evaluate the effectiveness of our proposed method for tree trunk segmentation and DBH calculation in the context of these prevalent tree species in an urban environment.

The data collection process involved capturing images of tree trunks using standard Android mobile phones, namely Galaxy S22 Ultra. These images were then used to develop and fine-tune the SegFormer model to accurately segment tree trunks. The resulting dataset of segmented images served as the foundation for our analysis and experimentation.

Including different species, such as Date palms, Vachellia nilotica, and Ziziphus mauritiana, enhances the diversity and applicability of our study. By considering multiple tree species commonly found in urban street environments, we can assess the generalizability and robustness of our proposed method across a range of tree types.

This dataset comprises a collection of paired images captured successively, encompassing both far and close images. The close images are obtained by carefully relocating the camera a specific distance closer to the tree from the original far image position. Rigorous attention is given to accurate measurements, including the ground truth distance from the far image to the tree, the incremental distance covered during image capture, and the precise diameter at breast height (DBH). These measurements help us assess how well our method separates tree trunks and calculates DBH. By using this dataset, we ensure the accuracy and reliability of our analysis.

The dataset consists of a total of 400 images collected using the Samsung Galaxy S22 Ultra smartphone. Sample images of both datasets are shown in Figure~\ref{sample_dataset_images}. We have used the RoboFlow \cite{Roboflow25:online} to annotate the data for the semantic segmentation task. 

\begin{figure*}[t]
\centering
\includegraphics[scale=0.33]{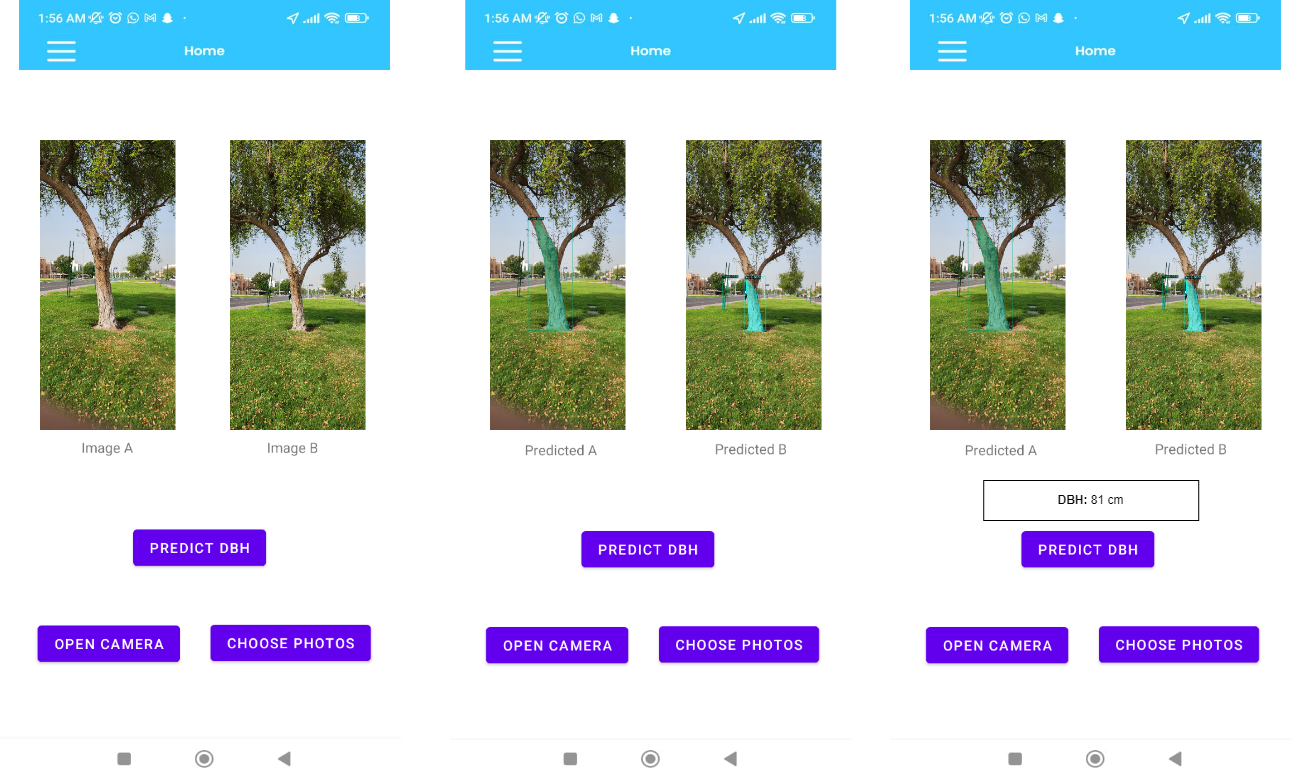}
\caption{Mobile Interface for DBH Estimation and Tree Trunk Segmentation:a) Image Pair Acquisition, b) Tree Trunk Segmentation with SegFormer \cite{NEURIPS2021_64f1f27b}, c) DBH Estimation and Display of Calculated Value}
\label{app123}
\end{figure*}

\subsection{Model Training}
\label{sec:discuss}

We have designed and implemented an Android application that performs various functions. Firstly, it captures images using the mobile's camera. These images are then transmitted to the cloud server hosted on AWS EC2 using web sockets. The purpose of sending the images to the server is to leverage a semantic segmentation model deployed on the server for inference, enabling the identification of tree trunks accurately. Figure~\ref{app123} illustrates the graphical user interface of the application, where users have the option to capture images or select them from their gallery. These chosen images are subsequently transmitted to a cloud server for further processing. The server utilizes the SegFormer model to extract the masked trunk output and estimates the Diameter at Breast Height (DBH) specifically for the identified tree.

In addition to the images, the application also transmits essential parameters such as sensor height, sensor width, focal length, and the distance covered during image capture. These parameters are crucial for conducting DBH estimation. By sending this information to the server, the necessary calculations for DBH estimation can be performed accurately.

In this research work, the authors utilized a deep-learning-based algorithm, the SegFormer \cite{NEURIPS2021_64f1f27b} model, for the semantic segmentation task.

Firstly, the dataset is pre-processed by resizing the images to a consistent resolution and normalizing the pixel values. The dataset is then split into training, validation, and testing splits. The number of images used in each split is shown in Table~\ref{tab:dataset-images}:

\begin{table}[t]
\centering
\begin{tabular}{cc}
\hline
\textbf{Dataset} & \textbf{Number of Images} \\ \hline
Train & 300 \\
Validation & 50 \\
Test & 50 \\ \hline
\end{tabular}
\caption{Distribution of Dataset for Training, Validation, and Test Images}
\label{tab:dataset-images}
\end{table}

% \begin{figure*}[h]
% \label{Fig-7}
% \begin{center}
% \includegraphics[width=\textwidth]{images/App-Updated.png}
% % \includegraphics[scale=0.35]{images/App-Updated.png}
% \end{center}

% \end{figure*}

For model initialization, a pre-trained backbone network, i.e. ResNet, is selected, and thus loaded with the corresponding weights. To proceed with the training process, the parameters such as the learning rate, batch size, and the number of training epochs are initialized as 0.01, 16, and 150. In this approach, we did not apply any augmentation techniques, as the tree trunk has to be in a straight orientation.

The model is been trained on the given dataset by iterating over the training data and feeding the images into the model. The model was trained on a computer with an Intel Core i7-13700K @2.5 GHz processor, 64 GB of RAM, and an Nvidia GeForce RTX 4090 graphics card. 

The model's weights were optimized using a suitable loss function, such as pixel-wise cross-entropy and the Dice coefficient, to measure the discrepancy between the predicted and ground truth segmentation. The model's performance was monitored on the validation set throughout the training process, assessing metrics such as mean Intersection over Union (mIoU) and pixel accuracy. The evaluation metrics of the model are given in Table~\ref{SegRes}.

After the completion of training, the model was fine-tuned by adjusting hyper-parameters, such as the learning rate and regularization techniques, to refine its performance further. Finally, the trained SegFormer model was evaluated on a separate test dataset to measure its segmentation accuracy and generalization capability.

%%%%%%%%%%%%%%%%%%%%%%%%%%%%%%%%%

\begin{figure}[h]
\begin{center}
\includegraphics[scale=0.35]{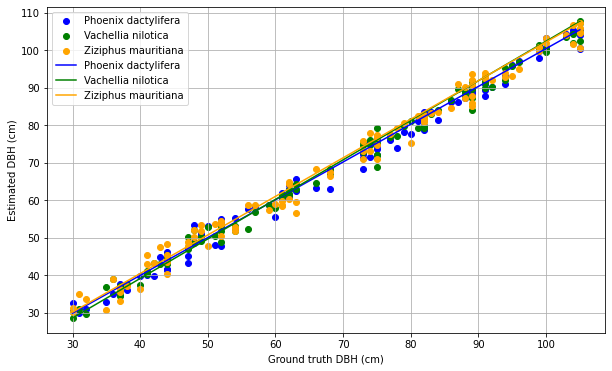}
\end{center}
\caption{Visual comparison of ground truth and estimated DBH}
\label{GTvsEst_DBH}
\end{figure}

\begin{figure}[h]
\begin{center}
\includegraphics[scale=0.35]{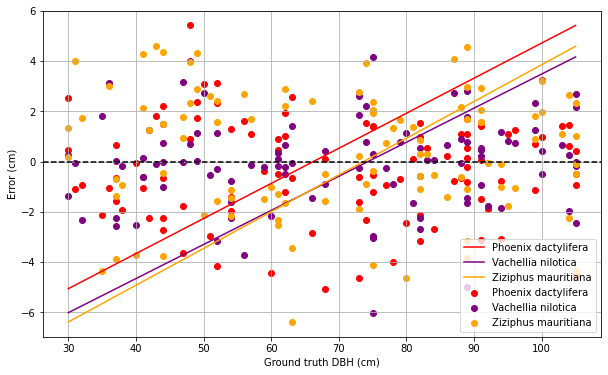}
\end{center}
\caption{Error Analysis of Estimated and Ground-Truth DBH for Each Tree Species}
\label{Error_Species}
\end{figure}

%%%%%%%%%%%%%%%%%%%%%%%%%%%%%%%%%%%%%%%%%%%%%%%%%%%%%%%%%%%%%%%%

\BB{\section{Application Inference}
In this section, we have introduced critical steps for users to follow when utilizing the application. Specifically, we have employed a fixed distance-based approach, utilizing distances of 20 and 15 feet for far and close images, respectively. The camera is positioned at breast height to ensure consistent image capture conditions. Distance measurement is conducted manually to attain precise distance values, which enhances result accuracy. However, as part of future work, we propose implementing an automated process based on a two-image approach. This automated process would calculate the estimated distance, enabling subsequent calculations based on this precise distance.

For the inference phase, we acquire two images (one for each far and close image) through the application and send them to the Amazon Elastic Compute Cloud (EC2) server via web sockets. This transmission includes essential camera characteristics such as focal length, image size, and sensor dimensions. The server on EC2 processes the images by performing mask predictions corresponding to target trees, applying experimental calculations, and estimating the Diameter at Breast Height (DBH). Subsequently, the server returns the processed images with masks, followed by the estimated DBH value. The entire process, from sending an image to retrieving the estimated DBH value, typically takes 7 to 9 seconds. The duration may vary depending on the mobile phone's connectivity, as larger image dimensions may be involved.

When trees do not typically share a crown at breast height due to a distinct spatial separation, special considerations arise when they exhibit crown overlap. In such instances, two approaches can be employed to address this scenario. Firstly, it is essential to position the target tree precisely at the centre of the captured image. This positioning ensures that the algorithm focuses solely on the target tree, excluding the neighbouring tree from consideration. Secondly, the user must carefully adjust their proximity to the tree in question to prevent crown interference at the breast height, as this particular height is of paramount importance in the analysis. 

}

\begin{table}[htbp]
\centering
\caption{Performance evaluation of SegFormer \cite{NEURIPS2021_64f1f27b} on Tree Trunk Segmentation}
\begin{tabular}{lc}
\toprule
\label{SegRes}
\textbf{Metric} & \textbf{Value} \\
\midrule
IoU (\%) & 87.4 \\
Pixel Accuracy (\%) & 93.8 \\
Mean Dice Coefficient & 0.91 \\
\bottomrule
\end{tabular}
\end{table}

%%%%%%%%%%%%%%%%%%%%%%%%%%%%%%%%%%%%%%%%%%%%%%%%%%%%%%%%%%%%%%%%%%%%%
\section{Results Analysis}
\label{sec:results}

\begin{table*}[t]
\centering
\caption{Evaluation metrics equations used for DBH}
\adjustbox{max width=\textwidth}{
\begin{tabular}{ccccc}
\hline
\toprule
\BB{\textbf{\text{RMSE (cm)}}} & \BB{\textbf{MAE (cm)}} & \BB{\textbf{reBias (\%)}} & \BB{\textbf{reRMSE (\%)}} & \BB{\textbf{Std. Dev. (cm)}} \\ \hline
\addlinespace
$\sqrt{\frac{1}{N}\sum_{i=1}^{N}(y_i - \hat{y}_i)^2}$ & $\frac{1}{N}\sum{i=1}^{N}|y_i - \hat{y}i|$ & $\frac{1}{N}\sum{i=1}^{N}\left(\frac{y_i - \hat{y}i}{y_i}\right)$  & $\sqrt{\frac{1}{N}\sum{i=1}^{N}\left(\frac{y_i - \hat{y}i}{y_i}\right)^2}$ & $\sqrt{\frac{1}{N}\sum{i=1}^{N}(x_i - \bar{x})^2}$ 
\\ 
\addlinespace
\hline
\end{tabular}
\label{Evaluation-metrics}
}
\end{table*}

\begin{table*}[t]
\centering
\caption{Statistical Results for three Tree Species namely  Date species (Phoenix dactylifera), Vachellia nilotica and Ziziphus mauritiana.}
\adjustbox{max width=\textwidth}{
\begin{tabular}{lccccccc}
\toprule
\label{DBHRes}
\textbf{Tree Species} & \BB{\textbf{RMSE (cm)}} & \BB{\textbf{MAE (cm)}} & \BB{\textbf{reBias (\%)}} & \BB{\textbf{reRMSE (\%)}} & \BB{\textbf{Min Error (cm)}} & \BB{\textbf{Max Error (cm)}} & \BB{\textbf{Std. Dev. (cm)}} \\
\midrule
Phoenix dactylifera        & 2.1                & 1.7               & -0.5                 & 4.8                  & -3.5                    & 4.2                     & 1.8                     \\
Vachellia nilotica        & 1.9                & 1.4               & 0.3                  & 3.6                  & -2.1                    & 3.7                     & 1.2                     \\
Ziziphus mauritiana        & 2.3                & 1.9               & -0.2                 & 5.0                  & -3.1                    & 4.8                     & 2.0                     \\
\midrule
\textbf{Average}             & 2.1                & 1.6               & -0.1                 & 4.5                  & -2.9                    & 4.2                     & 1.6                     \\
\bottomrule
\end{tabular}
}
\end{table*}

For evaluation of both distance and DBH calculations, the metrics which include root mean square error (RMSE), mean absolute error (MAE), relative Bias (reBias), relative RMSE (reRMSE), minimum error, maximum error, and standard deviation are used to verify the precision of our technique. The equations for these evaluation metrics are shown in Table~\ref{Evaluation-metrics}. \BB{Here, RMSE measures the average magnitude of prediction errors, offering insights into overall accuracy. MAE calculates average absolute differences between predicted and actual values. The reBias expresses bias as a percentage relative to actual values, indicating overestimation or underestimation tendencies. The reRMSE represents RMSE as a percentage relative to the average of actual values, aiding error assessment about the data scale. Standard Deviation characterizes data spread around the mean, revealing insights into data distribution and consistency.}

\begin{table*}[htbp]
\centering
\caption{A comparative analysis of different studies for measuring tree diameter at breast height (DBH). %\cite{SHAO2022107140}. 
Here the following terminologies are used: LCIS: Low-Cost Integrated sensor, \BB{SITDE: Single-Image Tree Diameter Estimation}, \BB{FS: Forest Scanner}, \BB{LiS: LiDAR-equipped smartphone}. The Range refers to the size of the DBH measured. N provides the count for the total number of trees used in the experiment. The forest type ranges from AF: artificial forest, MF: montane Forest, UAF: urban artificial forest, \BB{LC: Laurel Creek, BW: Beachwoods, VC: Van Cortlandt}, \BB{Con: Conifer, BL: Boradleaf}, \BB{Dec: Deciduous, and Conif: Coniferous}. Whereas the main species used in these studies are: MS0: Phoenix dactylifera, Vachellia nilotica and Ziziphus mauritiana MS1: Italian poplar, Camphor tree, Goldenrain tree, Arborvitae and Populus Canadensis, Pine; MS2: \BB{broadleaf, conifer, beech, black oak, red maple}; MS3: \BB{dwarf bamboos}; MS4: \BB{Calabrian pine, oriental plane}. Lastly, the price column refers to the device's total cost in each experiment study.}

\label{tab:method_comparison}
\adjustbox{max width=\textwidth}{
\begin{tabular}{ccccccccccc}
\hline
\toprule
\textbf{Method} & \textbf{Bias (cm)} & \textbf{reBias (\%)} & \textbf{RMSE (cm)} & \textbf{reRMSE (\%)} & \textbf{Range (cm)} & \textbf{N} & \textbf{Forest type} & \textbf{Main species} & \textbf{Price} \\
\hline
\midrule
\textbf{Ours} & $-0.05$ & $-0.1$ & $2.1$ & $4.5$ & $(30, 100)$ & $400$ & Urban Streets & MS0 & \$\textbf{300} & This study \\

LCIS & $-0.11$ & $-0.23$ & $1.55$ & $6.64$ & $(6, 51)$ & $371$ & AF/MF/UAF & MS1 & \$255 & \cite{SHAO2022107140} \\

\BB{SITDE} & $0.6$ & $-$ & $3.70$ & $-$ & $(6, 105)$ & $97$ & \BB{LC/BW/VC} & MS2 & \$600 & \cite{rs15030772} \\

\BB{FS} & $0.2$ & $0.75$ & $2.30$ & $10.4$ & NA & $672$ & \BB{Con/BL} & MS3 & $>$\$749 & \cite{tatsumi2023forestscanner}\\

\BB{LiS} & $-$ & $-5.10$ & $2.33$ & $-$ & $(8.5 , 37.5)$ & $105$ & Dec/Conif & MS4 & $>$\$700 & \cite{gulci2023measuring} \\
% SfM & $-0.77$ & $-3.77$ & $1.62$ & $7.73$ & $(4, 64)$ & $140$ & UAF & MS4 & $>$\$450 & \cite{f10080701} \\
% TOF & $0.33$ & $1.78$ & $1.26$ & $6.39$ & $(5, 40)$ & $193$ & AF & MS5 & \$600 & \cite{rs10111845} \\
\hline
\end{tabular}
}
\end{table*}

\BB{
Table~\ref{DBHRes} presents statistical measurements for tree species named Phoenix dactylifera, Vachellia nilotica, and Ziziphus mauritiana. For Phoenix dactylifera, the statistical analysis yielded an RMSE of 2.1 cm and an MAE of 1.7 cm, indicating the average differences between predicted and actual DBH values. The reBias was estimated at -0.5\%, suggesting a slight underestimation in the predictions compared to actual values. Additionally, the relative RMSE (reRMSE) was calculated as 4.8\%.

In the case of Vachellia nilotica, the statistical measurements revealed an RMSE of 1.9 cm and an MAE of 1.4 cm, signifying the average differences between predicted and actual DBH values. The reBias was estimated at 0.3\%, indicating a slight overestimation in the predictions, while the reRMSE was 3.6\%.

For Ziziphus mauritiana, the statistical analysis showed an RMSE of 2.3 cm and an MAE of 1.9 cm, representing the average differences between predicted and actual DBH values. The reBias was estimated at -0.2\%, indicating a slight underestimation in the predictions. Additionally, the reRMSE was calculated as 5.0\%.

}

% Additionally, the table also provides the average values across all the tree species. These average values represent the overall performance of the predictions in terms of RMSE, MAE, reBias, reRMSE, minimum error, maximum error, and Standard Deviation.

The provided statistical results offer a comprehensive evaluation of the accuracy, bias, and variability in DBH estimation for various tree species using an Android phone. These metrics provide valuable insights into the predictive model's performance for each species and the overall performance across all species. In comparison to the existing methods listed in Table \ref{tab:method_comparison}, our proposed method demonstrates exceptional accuracy, characterized by minimal negative bias and a relative bias of -0.05 cm and -0.1\%, respectively. It also achieves a low RMSE of 2.1 cm and a reRMSE of 4.5\% excluding the LCIS approach. These results signify precise and consistent forest inventory estimations, making our approach particularly well-suited for urban street environments and main species MS0 all at a competitive price of \$300. Contrasting with alternative methods like LCIS, which exhibit slightly higher biases and lower RMSE values, underscores the importance of selecting the most suitable approach based on specific forest types, main species, and budget constraints. Our method guarantees reliable and cost-effective forest inventory estimation, making it a compelling option in this field, providing both affordability and efficient performance.

Figure~\ref{GTvsEst_DBH} illustrates the distribution of tree DBH values compared to their corresponding ground truth values. The graph reveals that most trees exhibit DBH measurements ranging from 30 to 55 cm. This information provides insights into the typical DBH range for the trees in question. Additionally, Figure~\ref{Error_Species} presents a visual analysis of the RMSE for each tree species, compared to their respective ground truth DBH values. This analysis aids in understanding the accuracy of the DBH estimation for each species, providing valuable information on the performance of the DBH prediction model. 

%%%%%%%%%%%%%%%%%%%%%%%%%%%%%%%%%%%%%%%%%%%%%%%%%%%%%%%%%%%%%%%%%%%%

\section{Conclusion}
\label{sec:coclusion}

In conclusion, our study has successfully developed an innovative system for tree trunk segmentation and Diameter at Breast Height (DBH) estimation using images captured from Android mobile phones. Our approach eliminates the need for specialized equipment and training, making it a cost-effective and accessible solution for tree inventory.

Our research specifically focused on three tree species, namely Phoenix dactylifera, Vachellia nilotica, and Ziziphus mauritiana. The performance metrics indicate promising results, with Phoenix dactylifera showing an RMSE of 2.1 cm and a MAE of 1.7 cm, Vachellia nilotica exhibiting an RMSE of 1.9 cm and a MAE of 1.4 cm, and Ziziphus mauritiana demonstrating an RMSE of 2.3 cm and a MAE of 1.9 cm.

By leveraging image processing, cloud computing, and mobile technologies, our approach has the potential to revolutionize urban forestry management. It offers a practical and precise method for measuring tree trunks, generating comprehensive inventories, and utilizing this information to inform resource management, environmental protection, and urban planning decisions.

Moving forward, we plan to improve the system further and collaborate with stakeholders to implement it on a larger scale. We anticipate that this will significantly impact urban forestry practices, promote sustainable management of urban green spaces, and contribute to the overall well-being of cities.
%%%%%%%%%%%%%%%%%%%%%%%%%%%%%%%%%%%%%%%%%%%%%%%%%%%%%%%%%%%%%%%%%%%%
% -------------------------------------------------------------------------
\section*{Acknowledgment}

This work was supported in part by the Khalifa University of Science and
Technology under Faculty Start-Up under Grants FSU-2022-003 and 84740 0
0401.

%\bibliographystyle{ieeetr}
%\bibliography{ieeeRef.bib}

%%%%%%%%%%%%%%%%%%%%%%%%%%%%%%%%%%%%%%%%%%%%%%%%%%%%%%%%%%%%%

\end{document}